\documentclass[11pt]{article}

\usepackage{acl}

\usepackage{times}
\usepackage{latexsym}

\usepackage[T1]{fontenc}

\usepackage[utf8]{inputenc}

\usepackage{microtype}

\usepackage{inconsolata}

\usepackage{graphicx}

\AtBeginDocument{
  \setlength{\abovedisplayskip}{3pt}
  \setlength{\belowdisplayskip}{3pt}
  \setlength{\abovedisplayshortskip}{0pt}
  \setlength{\belowdisplayshortskip}{3pt}
}

\usepackage{amsmath,amssymb,amsfonts,amsthm}
\usepackage{verbatim} 
\usepackage{booktabs}
\usepackage{subcaption}
\usepackage{multirow}

\usepackage{algorithm}
\usepackage{algorithmic}

\usepackage{xcolor}

\usepackage{enumitem}

\newcommand\nprobe[0]{$n_{\text{probe}}$}

\title{Flash-GMM: A Memory-Efficient Kernel for Scalable Soft Clustering}

\author{
Gal Bloch \and Ariel Gera \and Matan Orbach \and Ohad Eytan \and Assaf Toledo \\
IBM Research \\
\texttt{gal.bloch@ibm.com} \\
\url{https://github.com/IBM/Flash-GMM} \\
}

\begin{document}
\maketitle

\begin{abstract}

We present \textbf{Flash-GMM}, a fused Triton kernel for efficient computation of 
Gaussian Mixture Models (GMMs) 
over large-scale data in a single GPU pass.
By eliminating the need to materialize the full responsibility matrix in GPU memory, Flash-GMM achieves a \textbf{20$\times$} speedup over existing implementations and enables training on datasets more than \textbf{100$\times$} larger than previously feasible on one device.
To demonstrate its 
impact, we 
integrate 
Flash-GMM into the IVF coarse quantizer for approximate nearest-neighbor (ANN) search.
We show that soft GMM clustering is now a viable drop-in replacement for $k$-means, and that GMM responsibilities can be leveraged to assign border vectors to multiple clusters.
Our approach reaches fixed recall targets with up to $1.7\times$ fewer distance computations, or equivalently, yields $+2$--$12$ 
recall@10 at matched computational cost.  
We release the kernel as an open-source project.
\end{abstract}

\section{Introduction}

Gaussian Mixture Models (GMMs) are versatile probabilistic tools that have found wide application across domains ranging from computer vision to bioinformatics. 
GMMs fit a statistical model to the data by estimating its underlying probability density as a mixture of Gaussians.

\begin{table}[ht]
\centering
\small
\begin{tabular}{rrrr}
\toprule
$N$ & Flash-GMM & vs.\ SciPy & vs.\ TorchGMM \\
\midrule
10K   & \textbf{85\,ms}       & 766$\times$     & 32$\times$ \\
50K   & \textbf{261\,ms}      & 1{,}260$\times$ & 20$\times$ \\
100K  & \textbf{444\,ms}      & 1{,}458$\times$ & 23$\times$ \\
250K  & \textbf{1{,}032\,ms}  & 1{,}597$\times$ & 19$\times$ \\
500K  & \textbf{2{,}069\,ms}  & 1{,}571$\times$ & 20$\times$ \\
1M    & \textbf{3{,}755\,ms}  & 1{,}738$\times$ & 22$\times$ \\
10M   & \textbf{7{,}400\,ms}  & 1{,}740$\times$ & OOM        \\
50M   & \textbf{35{,}510\,ms} & 1{,}752$\times$ & OOM        \\
100M  & \textbf{74{,}270\,ms} & 1{,}782$\times$ & OOM        \\
\bottomrule
\end{tabular}
\caption{Runtime of 30 GMM EM iterations for different data scales ($N$),
with $K=1024$, $D=128$, on an A100-80\,GB GPU 
(OOM = out of GPU memory).
  Flash-GMM obtains significant speedup in comparison to
  TorchGMM~\citep{torchgmm2023}, an existing 
  GPU kernel, 
  and a CPU-based implementation from SciPy.
  }
\label{tab:kernel}
\end{table}


GMMs quantify the probabilities of assigning each data point to each of the mixture components. These probabilities, termed \textit{responsibilities}, are typically realized in a full matrix of size $N \times K$, where $N$ is the number of data points and $K$ is the number of Gaussian components. During parameter estimation via Expectation Maximization (EM, ~\citealp{dempster1977em}), this matrix is recomputed at every iteration.

The massive parallelism offered by GPUs allows significantly faster GMM estimation. However, as GPU memory is limited, materializing the full responsibility matrix becomes impractical even for moderately sized datasets. Existing GPU implementations such as TorchGMM~\citep{torchgmm2023} run out of memory beyond $10$ million
data points, while CPU-based solvers~\citep{2020SciPy-NMeth} are orders of magnitude too slow (see Table~\ref{tab:kernel}). Thus, large scale applications based on GMM have remained out of reach at production scales.

To address this gap, we introduce \textbf{Flash-GMM}, a fused Triton~\citep{tillet2019triton} kernel that 
performs GPU-accelerated GMM estimation
without materializing the responsibility matrix in the GPU HBM memory. The design is inspired by the IO-aware tiling strategy of FlashAttention~\citep{dao2022flashattention}, adapted to the EM algorithm. The resulting kernel requires only $O(KD)$ GPU memory, where $D$ is the data dimensionality. Because memory usage does not grow with dataset size $N$, Flash-GMM can handle arbitrarily large datasets. Moreover, by performing all tile-local computations in GPU registers rather than round-tripping through GPU main memory, the kernel minimizes main memory access, thus alleviating the primary latency bottleneck.
Empirically, our kernel achieves a \textbf{20$\times$} speedup over existing implementations and enables training on datasets more than \textbf{100$\times$} larger than previously feasible on one device (see Table~\ref{tab:kernel}).

Our contribution of an efficient GMM kernel opens the door to many practical use cases. We demonstrate this by focusing on a prominent application: the IVF index coarse quantizer~\citep{jegou2011searching}, commonly used for approximate nearest neighbor (ANN) search. The quantizer is typically implemented via $k$-means clustering~\citep{lloyd1982least}; instead, we propose Flash-GMM as a practical and performant alternative (\S\ref{sec:casestudy}).

Using Flash-GMM confers several advantages in the IVF setting. 
First, it enables fast IVF index construction, 
for data scales that are too compute-intensive for a CPU and were previously too memory-demanding for a single GPU. 
Second, with GMM, the estimated responsibilities (posterior probabilities of data points given Gaussian components) naturally yield soft assignments of vectors to clusters. 
This enables a \textit{multiple assignment} scheme (\S\ref{sec:multiassign}), such
that vectors near cluster boundaries can be associated with several clusters. This contrasts with the hard assignment of $k$-means, which forces such ambiguous boundary cases into a single cluster. 

For ANN search, this scheme directly translates to improved recall, as near-boundary vectors 
are no longer prematurely discarded from the search space.
We demonstrate this improvement through experiments on standard search benchmarks.

In summary, our contributions are:
\begin{itemize}[leftmargin=1em,topsep=2pt]
  \item We introduce the \textbf{Flash-GMM kernel}: a memory-efficient, IO-aware GPU kernel for GMM, enabling training on arbitrarily large datasets.
  \item We demonstrate the impact of Flash-GMM for a practical IVF application. In the IVF setup, the introduced kernel enables larger data scales, and, with the multi-assignment scheme, improves search quality and cost tradeoffs. Flash-GMM with multi-assignment delivers up to $1.7\times$ fewer distance computations at fixed recall and $+2$--$12\,$ recall at any matched compute budget (\S\ref{sec:results}).
  
\end{itemize}

\section{Gaussian Mixture Models}
\label{sec:gmm}

Formally, given a data matrix $\mathbf{X} = \{\mathbf{x}_1, \mathbf{x}_2, \dots, \mathbf{x}_N\}$ with $\mathbf{x}_i \in \mathbb{R}^D$,
GMMs model the data as generated from a mixture of $K$ Gaussian distributions \citep{bishop2006pattern}.\footnote{Here we restrict ourselves to isotropic Gaussian components, i.e., each covariance matrix is constrained to the form $\sigma_k^2 I$. This assumption ensures statistical stability while remaining computationally tractable (see Appendix~\ref{sec:appendix-isotropic}).} 
The probability density function of a sample $\mathbf{x}_i$ is defined as
\[
p(\mathbf{x}_i \mid \Theta) = \sum_{k=1}^{K} \pi_k \, \mathcal{N}(\mathbf{x}_i \mid \boldsymbol{\mu}_k, \sigma_k^2 \mathbf{I}),
\]
where $\pi_k$ denotes the mixture weight of component $k$, satisfying
\[
\sum_{k=1}^{K} \pi_k = 1, \qquad \pi_k \geq 0,
\]
$\mathcal{N}(\mathbf{x}_i \mid \boldsymbol{\mu}_k, \sigma_k^2 \mathbf{I})$ is the Gaussian distribution parameterized by mean vector $\boldsymbol{\mu}_k \in \mathbb{R}^D$ and isotropic variance $\sigma_k^2 \in \mathbb{R}$, and
$\Theta = \{\pi_k, \boldsymbol{\mu}_k, \sigma_k^2\}_{k=1}^{K}$ denotes all model parameters.
Parameter estimation is performed by maximizing the log-likelihood of the observed dataset:
\[
\log p(\mathbf{X} \mid \Theta)
=
\sum_{i=1}^{N}
\log
\left(
\sum_{k=1}^{K}
\pi_k \, \mathcal{N}_{ik}
\right),
\]
where
$
\mathcal{N}_{ik}
=
\mathcal{N}(\mathbf{x}_i \mid \boldsymbol{\mu}_k, \sigma_k^2 \mathbf{I}).
$

Typically, GMM parameter estimation is carried out using the Expectation-Maximization (EM) algorithm \citep{dempster1977em}, which alternates between an Expectation (E) step and a Maximization (M) step.
The E-step computes the posterior probability $r_{ik}$ of component $k$ and each sample $\mathbf{x}_i$, also referred to as the \textit{responsibility}.
First, the unnormalized assignment score for $k$ is defined as
\[
z_{ik}
=
\pi_k \, \mathcal{N}(\mathbf{x}_i \mid \boldsymbol{\mu}_k, \sigma_k^2 \mathbf{I}).
\]
The responsibility $r_{ik}$ is then obtained by normalizing these scores across all components:
\[
r_{ik}
=
p(k \mid \mathbf{x}_i, \Theta)
=
\frac{
z_{ik}
}{
\sum_{j=1}^{K}
z_{ij}
}.
\]
Intuitively, $r_{ik}$ measures the degree to which sample $\mathbf{x}_i$ belongs to cluster $k$, where
\[
0 \leq r_{ik} \leq 1,
\qquad
\sum_{k=1}^{K} r_{ik} = 1.
\]
To improve numerical stability, responsibilities are computed in log-space using the log-sum-exp trick \citep{bishop2006pattern,blanchard2021accurately}.

In the M-step, the model parameters are updated using the responsibilities computed in the E-step. First, the effective number of samples assigned to component $k$ is computed as
\begin{equation}
N_k
=
\sum_{i=1}^{N}
r_{ik},
\label{eq:Nk}
\end{equation}
and the mixture weights are set to
\[
\pi_k
=
\frac{N_k}{N}.
\]
The updated mean vector of $k$ is computed from the responsibilities by
\begin{equation}
\boldsymbol{\mu}_k
=
\frac{1}{N_k}
\sum_{i=1}^{N}
r_{ik} \mathbf{x}_i,
\label{eq:muk}
\end{equation}
followed by an updated variance:
\begin{equation}
\sigma_k^2
=
\frac{1}{D N_k}
\sum_{i=1}^{N}
r_{ik}
\left\|\mathbf{x}_i - \boldsymbol{\mu}_k\right\|_2^2.
\label{eq:sig}
\end{equation}

The E-step and M-step are repeated iteratively until convergence, typically determined by a sufficiently small change in the log-likelihood between successive iterations or by reaching a predefined maximum number of iterations. 

\section{Flash-GMM}
\label{sec:method}
In principle, GMM is well suited for GPU acceleration due to the massive parallelism and computational throughput offered by modern GPUs. Since the responsibility $r_{ik}$ for each data point $x_i$ depends only on the shared model parameters $\{\pi_k, \boldsymbol{\mu}_k, \sigma_k^2\}$ and not on other data points, the E-step computations across all $N$ vectors are fully independent and can be executed in parallel.
However, efficiently mapping a CPU-oriented implementation to GPU is non-trivial, as memory efficiency can quickly become the primary bottleneck and limit the benefits of the available compute \citep{dao2022flashattention}.

Consider a moderate setup, with $N=10$M, $D=128$, $K=2{,}048$: at 4 bytes precision, 
the responsibility matrix $\mathbf{R} \in \mathbb{R}^{N \times K}$ alone 
occupies $10^7 \times 2{,}048 \times 4 \approx 80\text{GB}$ of memory. Storing the input $\mathbf{X} \in \mathbb{R}^{N \times D}$ adds roughly $5\text{GB}$. Together, this exceeds the capacity of many GPUs.
While 
the memory footprint grows linearly with $N$, $D$, and $K$, in practice, $N$ dominates: datasets often contain millions of vectors, making naive GPU implementations infeasible.

Secondly, consider the memory bandwidth requirements of the EM procedure described in Section \ref{sec:gmm}. The E-step requires a full read of $\mathbf{X}$ ($ND$ reads) and a full write of the responsibility matrix $\mathbf{R}$ ($NK$ writes). The M-step then performs three additional full reads of $\mathbf{R}$ in (\ref{eq:Nk}), (\ref{eq:muk}), and (\ref{eq:sig}), as well as two full reads of $\mathbf{X}$ in (\ref{eq:muk}) and (\ref{eq:sig}), respectively. 
Overall, the number of accesses is $\sim 3ND + 4NK$.
These repeated passes 
generate substantial memory traffic and significantly increase the pressure on HBM bandwidth.

Thus,
a straightforward GPU implementation of GMM is fundamentally constrained by memory capacity and memory bandwidth. This limits scalability and prevents efficient utilization of the available compute power.

\subsection{Flash-GMM kernel}
\label{sec:kernel}
To enable GPU-based GMM estimation at scale, 
we introduce a new tile-based memory-efficient kernel: \textbf{Flash-GMM}. The new kernel is aimed at efficiently utilizing the parallel computational capacity of a GPU, while minimizing GPU memory access, and keeping peak memory use constrained.
The implementation is inspired by the work of  \citet{dao2022flashattention} and \citet{flashkmeans}. 



We divide $\mathbf{X}$ into contiguous tiles
$\mathbf{X}_1, \ldots, \mathbf{X}_T$ of $B_N$ rows each.
For each tile $\mathbf{X}_t$, we perform two steps:

\begin{enumerate}[leftmargin=1.5em,topsep=2pt,itemsep=2pt]
\item Compute the log-likelihood $\log z_{ik}$ for the tile vectors
against the $K$ components, and the log-normalizer
$\log Z_i = \log \sum_k z_{ik}$ via the numerically stable online
log-sum-exp~\citep{blanchard2021accurately,dao2022flashattention}.

\item Compute the responsibilities $r_{ik}$ (using the $\log Z_i$ normalizers from above), and accumulate per-tile sufficient statistics $N_k(t) = \sum_i r_{ik}$,
$\boldsymbol{M}_k(t) = \sum_i r_{ik}\,\mathbf{x}_i$, and $Q_k(t) = \sum_i r_{ik}\,\|\mathbf{x}_i - \boldsymbol{\mu}_k\|^2$.
\end{enumerate}

Within each step, the $K$ components are processed sequentially in blocks of size $B_K$, since loading all component parameters at once exceeds on-chip
memory. 
The log normalizers $\log Z_i$ accumulated in Step~1, and the per-tile accumulators of Step~2 ($N_k(t)$, $\boldsymbol{M}_k(t)$ and $Q_k(t)$) are maintained in on-chip memory across the inner sweep. 

After all tiles are processed, the per-tile contributions are atomically
reduced into global accumulators $\mathbf{N}, \boldsymbol{M}, \mathbf{Q}$,
from which the M-step recovers $\pi_k, \boldsymbol{\mu}_k, \sigma_k^2$ via
Eqs.~(\ref{eq:Nk})--(\ref{eq:sig}). The complete two-loop structure is
given in Algorithm~\ref{alg:flash-gmm}.






\begin{algorithm*}[t]
\small
\caption{\textsc{Flash-GMM} (Single EM Iteration)}
\label{alg:flash-gmm}
\begin{algorithmic}[1]
\REQUIRE Data matrix $\mathbf{X} \in \mathbb{R}^{N \times D}$ in HBM, GMM parameters $\{\pi_k, \boldsymbol{\mu}_k, \sigma_k^2\}_{k=1}^K$ in HBM.
\STATE Set block sizes $B_N$ (tile size, e.g.\ 64), $B_K$ (component block, e.g.\ 16). 
\STATE Initialize accumulators $\mathbf{N} = \mathbf{0} \in \mathbb{R}^K$, $\boldsymbol{M} = \mathbf{0} \in \mathbb{R}^{K \times D}$, $\mathbf{Q} = \mathbf{0} \in \mathbb{R}^K$ in HBM.
\STATE Divide $\mathbf{X}$ into $T = \lceil N / B_N \rceil$ tiles $\mathbf{X}_1, \ldots, \mathbf{X}_T$ of size $B_N \times D$ each.
\FOR[Parallel across GPU blocks]{$1 \leq t \leq T$}
    \STATE Load $\mathbf{X}_t$ from HBM to on-chip registers. Initialize $\log \mathbf{Z} = -\infty \in \mathbb{R}^{B_N}$ on chip.
    \FOR[Step 1: Log-likelihoods and log-normalizers]{$j = 1$ to $\lceil K / B_K \rceil$}
        \STATE Load $\{\pi_k, \boldsymbol{\mu}_k, \sigma_k^2\}$ for $k \in [(j{-}1)B_K {+} 1,\; jB_K]$ from HBM to SRAM.
        \STATE On chip, compute $\log z_{ik} = \log \bigl(\pi_k \,\mathcal{N}(\mathbf{x}_i \mid \boldsymbol{\mu}_k,\, \sigma_k^2 \mathbf{I})\bigr) \in \mathbb{R}^{B_N \times B_K}$.
        \STATE On chip, accumulate $\log Z_i \leftarrow \log\!\bigl(\exp(\log Z_i) + \textstyle\sum_{k} \exp(\log z_{ik})\bigr) \in \mathbb{R}^{B_N}$.
    \ENDFOR
    \FOR[Step 2: Responsibilities and sufficient statistics ($\log \mathbf{Z}$ remains in registers)]{$j = 1$ to $\lceil K / B_K \rceil$}
        \STATE Load $\{\boldsymbol{\mu}_k, \sigma_k^2\}$ for $k \in [(j{-}1)B_K {+} 1,\; jB_K]$ from HBM to SRAM.
        \STATE On chip, recompute $\log z_{ik}$ (same as line 10).
        \STATE On chip, compute $r_{ik} = \exp(\log z_{ik} - \log Z_i) \in \mathbb{R}^{B_N \times B_K}$.
        \STATE On chip, accumulate $N_k(t) \mathrel{+}= \sum_{i} r_{ik}$, \; $\boldsymbol{M}_k(t) \mathrel{+}= \sum_{i} r_{ik}\, \mathbf{x}_i$, \; $Q_k(t) \mathrel{+}= \sum_{i} r_{ik}\, \|\mathbf{x}_i - \boldsymbol{\mu}_k\|^2$.
    \ENDFOR
    \STATE Atomically add $N_k(t)$, $\boldsymbol{M}_k(t)$, $Q_k(t)$ to global accumulators $\mathbf{N}$, $\boldsymbol{M}$, $\mathbf{Q}$ in HBM.
\ENDFOR
\STATE Compute $\pi_k^{\mathrm{new}} = N_k / \textstyle\sum_{k'} N_{k'}$, \quad $\boldsymbol{\mu}_k^{\mathrm{new}} = \boldsymbol{M}_k / N_k$, \quad $(\sigma_k^2)^{\mathrm{new}} = Q_k / (D \cdot N_k)$ \quad for all $k$.
\RETURN $\{\pi_k^{\mathrm{new}},\; \boldsymbol{\mu}_k^{\mathrm{new}},\; (\sigma_k^2)^{\mathrm{new}}\}_{k=1}^K$.
\end{algorithmic}
\end{algorithm*}

\subsection{Memory Efficiency}
\label{sec:memory}

\paragraph{Peak memory use}
A significant advantage of Flash-GMM is that it never materializes the $N \times K$ responsibility matrix in HBM memory, thus keeping peak memory consumption low.

Instead, the kernel maintains two kinds of on-chip state.
First, within each tile $\mathbf{X}_t$, the per-vector log-normalizers
$\log Z_i$
are kept on chip, and are never written back to HBM.
That makes them
immediately available for computing the $r_{ik}$ responsibilities in Step~2.
Then, the per-tile accumulators $N_k(t)$, $\boldsymbol{M}_k(t)$, and $Q_k(t)$ are directly computed from the responsibilities.
Overall, the HBM only stores the GMM parameters ($\mathcal{O}(KD)$ elements), and streaming tiles of the data matrix $\mathbf{X}$. 
Thus, the kernel scales to arbitrarily large datasets.

\paragraph{HBM bandwidth.}
A second advantage of the described kernel is the reduced number of HBM memory accesses.
The data $\mathbf{X}$, with $ND$ elements, is read once from the HBM. 
The $KD$ parameters $\{\pi_k, \boldsymbol{\mu}_k, \sigma_k^2\}$  are read twice, yielding 
$ND$ + $2KD$ reads in total.
Compared to the naive baseline of Section~\ref{sec:method} ($3ND + 4NK$ memory accesses), Flash-GMM eliminates the $O(NK)$ responsibility-matrix traffic entirely, reducing total HBM accesses to $\mathcal{O}(ND)$.
That is the primary source of Flash-GMM's speedup over existing kernels.

\subsection{Implementation Details}

We implemented the Flash-GMM kernel using Triton~\citep{tillet2019triton} and validated it on an NVIDIA A100 (80GB) GPU against a SciPy CPU reference. While the development was done on A100, the kernel has been validated to produce correct results on H100 and RTX5080 GPUs as well; the core algorithmic ideas are hardware-agnostic, and the implementation is straightforward to adapt to new architectures.



The kernel uses a 1-D grid with $\lceil N / B_N \rceil$ blocks, where the tile size is $B_N = 64$, so each block processes 64 input vectors. Within a block, the $K$ components are processed in chunks of $B_K = 16$, and input vectors of dimension $D$ are padded to $B_D = 128$. Each block contains 4 warps, and blocks are scheduled independently across streaming multiprocessors (SMs). For $N = 10^6$, this gives $\lceil N / B_N \rceil = \textbf{15{,}625}$ parallel blocks, providing enough work to fully occupy the 108 SMs of the A100.

\subsection{Evaluation}
\label{sec:runtime}
To empirically validate the benefits of the new kernel, we test its 
runtime and memory costs across multiple scales of data. We compare Flash-GMM to two contemporary baselines: the CPU implementation (on an AMD EPYC 7763 processor) from SciPy~\citep{2020SciPy-NMeth}, and TorchGMM~\citep{torchgmm2023}, a GPU GMM kernel. 

\paragraph{Runtime and Scale} Table~\ref{tab:kernel} 
depicts the runtime measurements.
Overall, Flash-GMM is $\mathbf{766}$--$\mathbf{1{,}740}\times$ faster than
SciPy and $\mathbf{19}$--$\mathbf{32}\times$ faster than TorchGMM
across dataset sizes.

Critically, TorchGMM runs out of memory at $N>10^6$, while Flash-GMM
scales up to $N=10^8$ on the same hardware, enabling soft GMM training on
datasets more than $100\times$ larger than previously feasible on a
single device. Flash-GMM 
thus unlocks
new use cases while delivering substantial speedups over existing implementations.

\paragraph{Peak memory footprint}
Table~\ref{tab:memory} reports the GPU memory allocated by the kernel
itself 
for Flash-GMM and TorchGMM, across
dataset sizes at $K=1024$, $D=128$.
Flash-GMM's kernel allocation grows as $O(N)$ via the $\log Z_i$ buffer
($N \times 4$ bytes) plus $O(KD)$ accumulators ($\approx 0.5$\,MB fixed),
totalling \textbf{4.5\,MB} at $N=10^6$.
TorchGMM materializes the full $N{\times}K$ responsibility matrix and
intermediate tensors, consuming \textbf{21\,GB} at the same scale ---
a \textbf{4{,}668$\times$} larger kernel footprint.

\begin{table}[t]
\centering
\small
\begin{tabular}{rrr}
\toprule
$N$ & Flash-GMM & TorchGMM \\
\midrule
10K   & \textbf{0.6\,MB}  & 229\,MB  \\
50K   & \textbf{0.7\,MB}  & 1{,}067\,MB  \\
100K  & \textbf{0.9\,MB}  & 2{,}113\,MB  \\
250K  & \textbf{1.5\,MB}  & 5{,}262\,MB  \\
500K  & \textbf{2.5\,MB}  & 10{,}514\,MB \\
1M    & \textbf{4.5\,MB}  & 21{,}006\,MB \\
\bottomrule
\end{tabular}
\caption{Kernel GPU memory (excluding input data $X$) for Flash-GMM
  vs.\ TorchGMM~\citep{torchgmm2023}, $K=1024$, $D=128$, A100-80\,GB.
  Flash-GMM allocates only $\log Z_i$ ($N{\times}4$ bytes) plus
  $O(KD)$ accumulators; TorchGMM materializes the full $N{\times}K$
  responsibility matrix, and exhausts memory for $N > 1M$.}
\label{tab:memory}
\end{table}

\section{Novel Usage of GMMs for IVF}
\label{sec:casestudy}
In this section, we introduce a novel use case for GMMs that is enabled by the Flash-GMM kernel. We apply Flash-GMM to the IVF index coarse quantizer \citep{jegou2011pq}, which typically relies on K-Means clustering,  as a drop-in (\S\ref{sec:dropin}) and with the addition of soft assignment (\S\ref{sec:multiassign}). 

Prior GMM implementations, whether CPU- or GPU-based, are generally too compute- and memory-intensive to scale to the regimes required for practical IVF training.
By introducing an optimized and scalable 
implementation, Flash-GMM allows for soft clustering within the IVF pipeline. 

This unlocks several advantages rooted in probabilistic soft assignments. First, the soft training procedure provides additional flexibility during clustering, allowing vectors to contribute to multiple clusters according to their posterior probabilities rather than enforcing hard assignments. Second, the probabilistic formulation naturally yields a principled multi-assignment strategy derived directly from the final responsibilities, which can improve recall by assigning vectors to multiple coarse partitions in a statistically-grounded manner.


\subsection{IVF Indexing}

An Inverted File (IVF) index partitions $\mathbf{X}$ into $K$ cells using a coarse quantizer. 
This is almost universally implemented with $k$-means clustering~\citep{johnson2021billion}, where the cells correspond to clusters induced by the learned centroids.

Each vector $x_i$ is assigned to a single cell and stored in the corresponding \emph{posting list}. At query time, a query vector $q$ is assigned to its $n_{\text{probe}}$ nearest centroids, and only the vectors contained in the associated posting lists are compared against $q$ using the exact distance metric. The top-$r$ nearest vectors among these candidates are then returned as the final retrieval results.

\subsection{GMM as a Drop-in Coarse Quantizer}
\label{sec:dropin}

Replacing $k$-means with Flash-GMM in the IVF coarse quantizer requires no modifications to either the index structure or the query pipeline. The IVF index consumes only the cluster centroids produced by the quantizer, and Flash-GMM outputs centroids in the same format as $k$-means. Consequently, the search algorithm remains entirely unchanged.

During index construction, each vector is assigned to the cluster with the highest responsibility, 
thereby collapsing the soft partition induced by GMM into the standard IVF posting-list structure.

\subsection{GMM Multi-Assignment}
\label{sec:multiassign}


Even with improved centroids from soft clustering, a vector near a Voronoi
boundary is still stored in only one posting list.
If a query falls on the other side of the boundary, the vector is
invisible.
The GMM training provides an immediate remedy: the final-iteration
responsibilities $r_{ik}$ directly quantify how much each vector
``belongs to'' each cluster.
A vector with $r_{ik} = 0.45$, $r_{ij} = 0.42$ near two cluster
boundaries clearly deserves to be indexed in both.


We assign each vector $x_i$ to at most two clusters: the top-$2$ clusters whose responsibilities satisfy $r_{ik} > \tau$ with $\tau = 1/K$. The threshold $\tau = 1/K$
corresponds to the uniform prior probability of a cluster, so a cluster is selected only if observing $x_i$ increases its posterior probability beyond the prior.

\subsection{Experimental Setup}
\label{sec:setup}

We compare $3$ IVF coarse quantizers: 

\begin{itemize}[leftmargin=1em,itemsep=0.5pt, topsep=1pt]
    \item \textbf{K-Means}: $100$ iterations of $k$-means from the FAISS library \citep{faiss}, serving as industry standard
  baseline. Single assignment.
  
\item \textbf{GMM single}: Flash-GMM with single assignment of each vector to a cluster.
Model estimation warm-starts with $10$ iterations of FAISS $k$-means, followed by $90$ Flash-GMM iterations. 

\item \textbf{GMM multi}: Flash-GMM with multi-assignment using the $\tau=1/K$ responsibility threshold. The model estimation process follows the procedure described for GMM single.

\end{itemize}

\noindent The used datasets are:



\textbf{SIFT1M} \citep{jegou2011pq}: $10^6$ SIFT descriptors,
$D=128$, 10K queries, standard ground truth.

\textbf{Deep10M}: first $10^7$ vectors of the deep-image-96 dataset \citep{babenko2016efficient}, $D=96$, 10K queries,
ground truth computed via brute-force on the full 10M subset.

\textbf{GloVe-100}: $1.18{\times}10^6$ GloVe word embeddings,
$D=100$, 10K queries, angular distance~\citep{pennington2014glove}.
Vectors are L2-normalised before clustering (as is standard for
angular benchmarks); ground truth recomputed on normalised vectors.

Index construction uses FAISS \texttt{IndexIVFFlat}; centroids are
injected directly without re-training. For multi-assignment, vectors are inserted into multiple posting lists
in the standard FAISS index, without modifying the search routine. All methods use $K=1024$ and the same random seeds.

\paragraph{Quality metric}
We evaluate the search quality with \emph{Recall@$10$} (\emph{$R@10$}): the fraction of queries for which the true nearest neighbor (determined by brute-force) appears in the top-10 returned results.

\paragraph{Search cost metric}
We use the number of distance computation operations (DCO) per query as the primary cost metric. DCO directly measures the amount of search work performed, and is largely independent of hardware and the particular choice of $K$. Comparing methods using only \nprobe{} can be misleading, since \nprobe{} does not account for posting-list length. This is particularly important under multi-assignment, where improved recall may partially arise from scanning more vectors per probe due to longer posting lists. We therefore report \emph{$R@10$} as a function of both \nprobe{} and DCO.

\subsection{Results - Quality vs. Cost}
\label{sec:results}


Table~\ref{tab:multiassign} depicts a representative result from the GloVe-100 dataset, illustrating the recall-computation trade-off for the different approaches.

We see that for a given choice of \nprobe{}, GMM-single slightly outperforms K-Means at matched DCO; GMM-multi achieves substantially higher recall, but at the cost of higher DCO: at \nprobe{}=$16$, GMM multi gains $+7.0$ pp over K-Means
($0.92$ vs.\ $0.85$) while DCO rises from $18.4K$ to $32.8K$.

Importantly, the DCO increase is more than offset when comparing across operating points.
In the GloVe-100 example, GMM-multi's recall of $0.92$ at \nprobe{}=$16$ outperforms K-Means at \nprobe{}=$32$ in terms of both recall and DCO. GMM-multi thus simultaneously delivers higher recall \emph{and} lower computation than the next reachable single-assignment operating point. Full results are in App. Table~\ref{tab:multiassign-full}.

\begin{figure}[!h]
  \centering
  \begin{subfigure}[b]{.92\linewidth}
    \includegraphics[width=\linewidth]{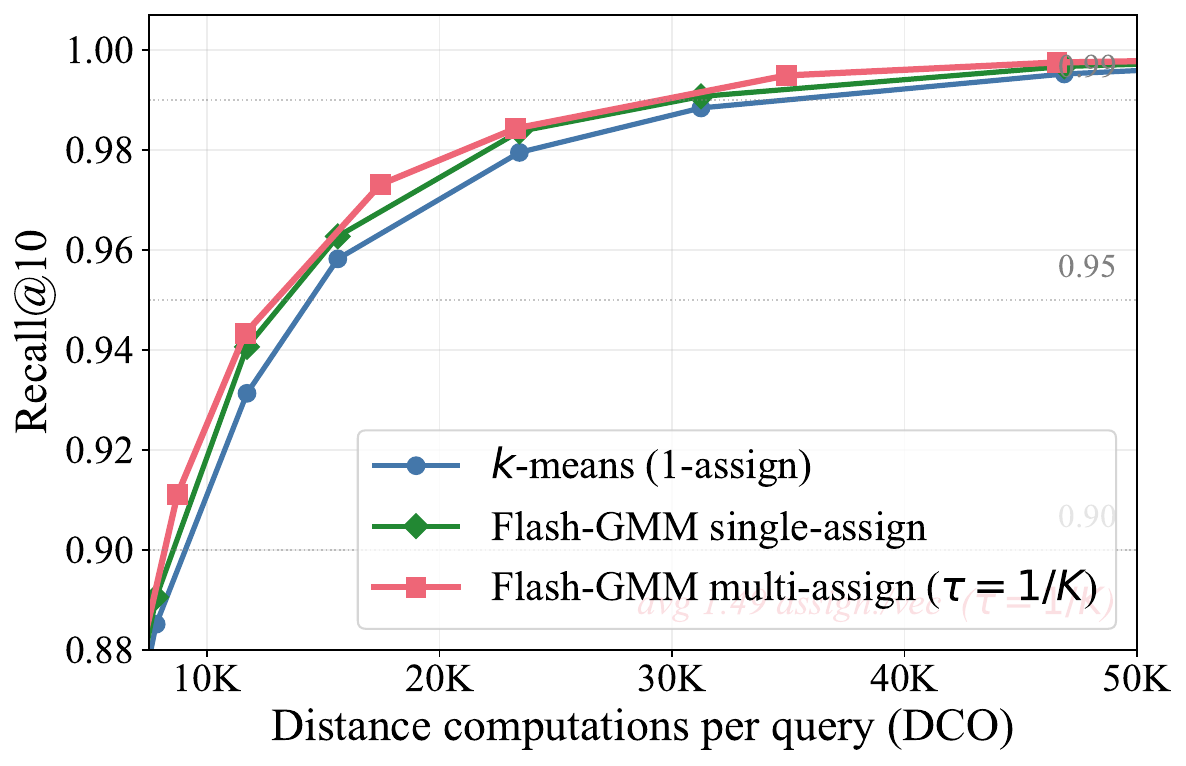}
    \caption{SIFT1M ($D=128$)}
  \end{subfigure}
  \hfill
  \begin{subfigure}[b]{.92\linewidth}
    \includegraphics[width=\linewidth]{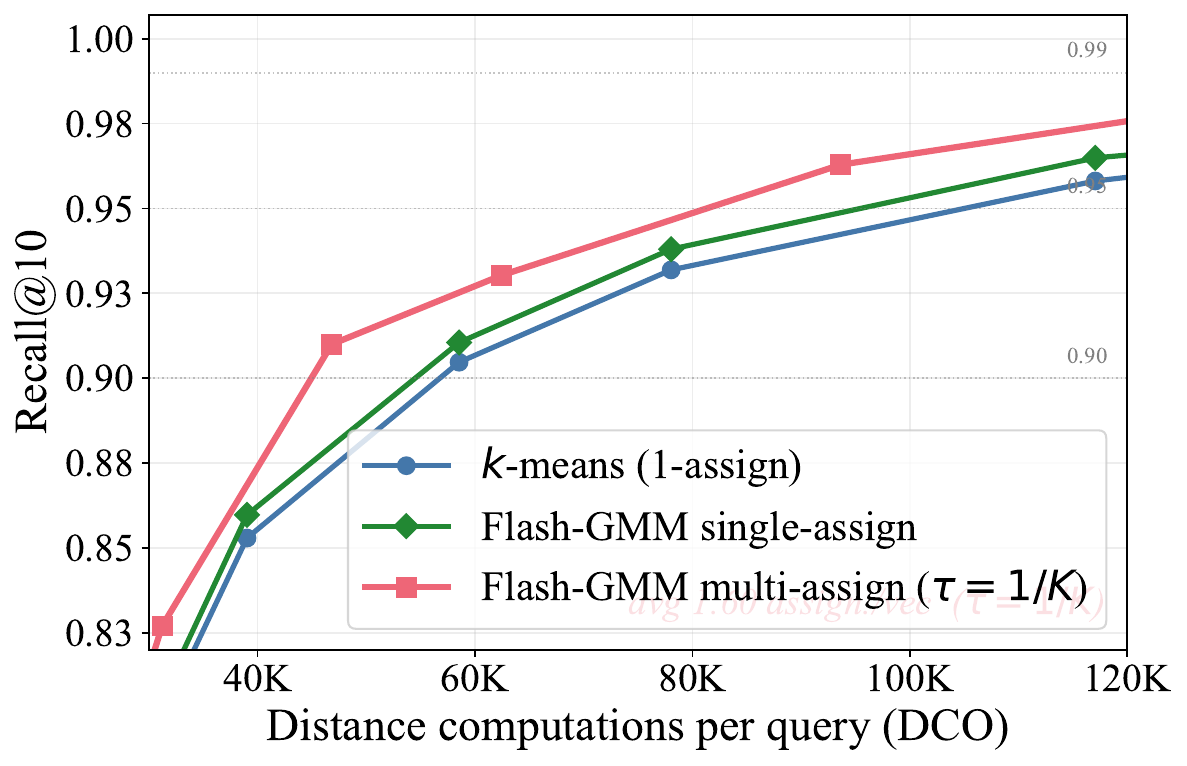}
    \caption{Deep10M ($D=96$)}
  \end{subfigure}
  \hfill
  \begin{subfigure}[b]{.92\linewidth}
    \includegraphics[width=\linewidth]{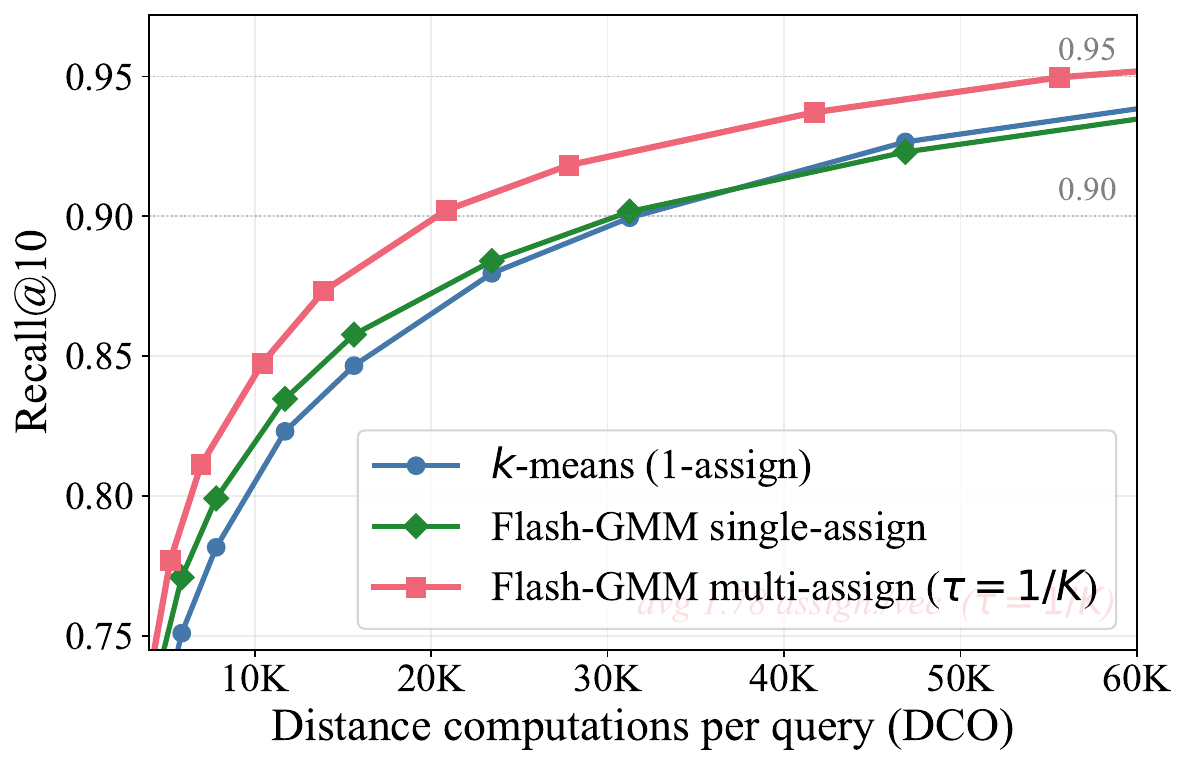}
    \caption{GloVe-100 ($D=100$)}
  \end{subfigure}
  \caption{Recall@10 vs.\ DCO ($K=1024$). Each point corresponds to one \nprobe{} value.
    GMM multi dominates both baselines
    across the full latency budget. 
    Dashed reference lines mark recall targets 0.90, 0.95, 0.99.
    }
  \label{fig:multiassign}
\end{figure}

Figure~\ref{fig:multiassign} depicts the full recall-DCO pareto curves, across all $3$ datasets. Each curve depicts six points, corresponding to the different \nprobe{} values. The figure demonstrates the consistent pareto-improvement pattern described above: because multi-assignment inflates list lengths, each GMM-multi point sits to the \emph{right} of the single-assignment point at the
\emph{same} \nprobe{} -- but it also sits \emph{above} the single-assignment point at an equivalent DCO, which corresponds to a single-assignment point with a \emph{higher} \nprobe{}. At every such comparison, GMM multi delivers strictly higher recall for the same compute budget. 

The gain is largest on GloVe (Figure~\ref{fig:multiassign}(c)), potentially since word-embedding spaces have
high \emph{semantic density}: many vectors cluster near boundaries between
topically related word clusters, yielding a high assignment average and more boundary vectors that benefit from multi-assignment.


\begin{table}[t]
\centering

\small
\setlength{\tabcolsep}{3pt}
\begin{tabular}{lcccccc}
\toprule
 & \multicolumn{2}{c}{K-Means} & \multicolumn{2}{c}{GMM single}
 & \multicolumn{2}{c}{GMM multi} \\
\cmidrule(lr){2-3}\cmidrule(lr){4-5}\cmidrule(lr){6-7}
np & R@10 & DCO & R@10 & DCO & R@10 & DCO \\
\midrule
16 & 0.85 & 18.4 & 0.86 & 18.4 & 0.92 & 32.8 \\
32 & 0.90 & 36.9 & 0.90 & 36.9 & 0.95 & 65.6 \\
\bottomrule
\end{tabular}

\caption{Recall@10 and DCO ($\times 10^3$) for GloVe-100 at selected \nprobe{} values, $K=1024$.
  For single-assignment methods DCO $= n_\text{probe} \times N/K$;
  for GMM multi, DCO $= n_\text{probe} \times N\bar{m}/K$
  ($\bar{m}=1.78$ for GloVe-100).
  Full results across all datasets and \nprobe{} values are in Appendix Table~\ref{tab:multiassign-full}.}
\label{tab:multiassign}

\end{table}

Experiments were executed with $3$ random seeds. Above we report the results for the first seed; we find that Recall@10 at any fixed \nprobe{} varies by less than $0.003$, and the multi-assignment recall gains exhibit similarly low variance. All reported results use $K=1024$, though we verified that the recall improvements persist across $K \in \{256, 1024, 4096\}$ (Appendix Table~\ref{tab:k-ablation}). All experiments use a warm-start initialized with 10 $k$-means iterations. We also evaluated $k$-means++ initialization~\cite{arthur2007kmeans}, but warm-start achieved comparable or better recall on most datasets while providing faster training (Appendix~\ref{sec:appendix-warmstart}).

\paragraph{Index build runtime}
\label{sec:trainingcost}
Table~\ref{tab:index_build_runtime} reports the index build wall-clock time for K-Means and Flash-GMM across all datasets, on a single Nvidia A100-80GB GPU.
Importantly, training a GMM coarse quantizer at $N=10^7$ (Deep10M) was previously impractical on an A100-80GB GPU: existing GMM implementations typically exhaust GPU memory beyond $N\!\approx\!10^6$. Scaling GMM training to this regime is thus a key contribution of Flash-GMM, enabled by its memory-efficient design.

In terms of runtime, Flash-GMM is approximately $2.5\times$--$3.3\times$ slower than K-Means. However, the absolute overhead amounts only to additional \emph{one-time} offline work.\footnote{Multi-assignment introduces 
no additional 
training cost beyond a single scan of the $N{\times}K$ responsibility matrix, which is negligible at $N=10^6$, $K=1024$.}
Moreover, the additional training cost is amortized over query time: multi-assignment reaches the same recall targets at substantially lower per-query DCO, so the savings compound across the lifetime of the index.


\begin{table}[t]
\centering
\resizebox{\columnwidth}{!}{
\begin{tabular}{lcccc}
\toprule
 & K-Means & Flash-GMM & TorchGMM & SciPy \\
\midrule
SIFT1M   & 3.5s  & 11.8s  & 247.8s   & $\approx$5.6h  \\
Deep10M  & 53.3s & 123.1s & OOM      & $\approx$58h   \\
GloVe-100 & 6.4s & 21.8s  & 493.86s  & $\approx$10h   \\
\bottomrule
\end{tabular}
}
\caption{Index build runtime for K-Means, Flash-GMM, TorchGMM, and SciPy. Due to prohibitively long
wall-clock times, SciPy runtimes are approximated based on the
speedup factors observed in Table~\ref{tab:kernel}.}
\label{tab:index_build_runtime}
\end{table}





\subsection{Multi-Assignment Analysis}

We analyze the practical behavior of the proposed multi-assignment strategy.

Under the chosen responsibility threshold $\tau = 1/K$, the average number of posting lists associated with a vector is $1.49$ for SIFT1M, $1.60$ for Deep10M, and $1.78$ for GloVe-100. In practice, most vectors have one dominant assignment, or at most two clusters whose responsibility exceeds $\tau$.

We additionally evaluated $\tau=2/K$ and $\tau=0.5/K$, and found both inferior to $\tau=1/K$. The former reduces recall by 1--4 percentage points at every \nprobe{}, while yielding only marginal search-time savings. The latter increases search time due to larger posting lists, without improving recall.

To isolate the benefit of GMM responsibilities from multi-assignment alone, we compare GMM multi against \textbf{K-Means hard top-2}. This baseline assigns each vector to the posting lists of its two nearest $k$-means centroids according to L2 distance, regardless of proximity to the cluster boundary.

Despite doubling the posting-list size, hard top-2 requires up to $\mathbf{1.8\times}$ more DCO than 
single-assignment $k$-means to achieve the same recall (Appendix~\ref{sec:appendix-hard-multi}). This indicates that the gains of GMM multi are not explained by multi-assignment alone. 

Prior work also observes this limitation of hard top-2 assignment. To obtain higher-quality multi-assignments for $k$-means-based indexing, RAIRS \citep{yang2026rairs} propose the AIR heuristic for determining the posting lists.
Since they did not release public code, we cannot directly compare to their approach; for completeness, we adapt GMM multi-assignment to their chosen index configuration, and show that GMM multi achieves competitive recall with less index inflation, compared to their reported results on a shared dataset (Appendix~\ref{sec:appendix-ivfpq}).

\section{Related Work}
\label{sec:related}

\paragraph{Soft-assignment quantizers.}
To our knowledge, no prior work applies scalable soft-assignment EM to
the IVF quantizer.

\paragraph{Fine-quantizer methods.}
Product quantization (PQ, \citealp{jegou2011pq}) and its extensions improve the
residual quantizer applied \emph{after} coarse assignment, and are
orthogonal to Flash-GMM.

\paragraph{Assignment-side methods.}
RAIRS~\citep{yang2026rairs} corrects boundary assignment errors by storing each vector in two posting lists using the AIR geometric heuristic to select the second list, without altering centroids --- assigning a second list to \emph{every} vector regardless of boundary proximity.


\paragraph{IO-aware kernels.}
Flash-$k$-means~\citep{flashkmeans} applies a fused distance kernel to reduce
HBM bandwidth in hard-assignment clustering, sharing the IO-aware tiling
motivation of our work.
Our kernel targets the GMM E-step: it computes soft responsibilities via a
numerically stable log-sum-exp and accumulates responsibility-weighted
sufficient statistics in the same pass --- operations with no analogue in
hard-assignment $k$-means.

\section{Discussion}
In this work we presented two complementary contributions to scalable soft clustering with GMMs.

\textbf{Flash-GMM} makes GMM training practical at scale, via a fused Triton kernel with $O(KD)$ working memory, eliminating the memory barrier that previously confined soft EM to small datasets.

\textbf{GMM multi-assignment} reuses the final Flash-GMM responsibilities to store boundary vectors in multiple posting lists.


Together, the two methods push the recall--compute Pareto frontier
significantly beyond what either achieves alone.
In our application on the IVF coarse quantizer, we demonstrate the practical utility unlocked by Flash-GMM multi-assign --- reaching superior results compared to $k$-means, while operating on large datasets for which GMM was not previously feasible.

The benefits we demonstrate in the context of IVF coarse quantization can potentially be coupled with other advances, most notably fine-quantizer methods such as IVF-PQ~\citep{jegou2011pq} and IVF-PQfs~\citep{andre2016cache}. Future work can explore the potential for such combinations to further promote ANN search applications. 

The IVF use case, however, is merely one example; GMMs have varied applications across multiple domains, which could similarly benefit from this efficient Flash-GMM implementation. In medical imaging, for instance, a probabilistic GMM approach has proven useful for image segmentation problems~\citep{song2014extension,riaz2020gaussian}. Similarly, GMMs are explored for diverse use cases in genomics, such as clustering gene expression patterns~\citep{liu2022gmmchi} or modeling relationships between the genomes of different organisms~\cite{clarke2018ggrasp}. Critically, many of the existing 
use cases for GMMs operate at particularly vast data scales, and are currently constrained to applying inefficient CPU-based implementations.

Furthermore, this unlocks the potential for increased use of GMMs. Prior work has shown GMMs can act as a drop-in replacement for $k$-means with improved partition quality across diverse workloads~\citep{PATEL2020158, liang2022gmmseg}; Flash-GMM removes the scalability barrier that has prevented this potential from being realized in practice.

The Flash-GMM kernel was optimized for the A100 GPU. The H100 architecture introduces 
new hardware primitives - most notably the Tensor Memory Accelerator and asynchronous warp-group MMA instructions. A Flash-GMM kernel redesigned around these primitives could exploit 
H100-specific features to deliver speedups beyond those reported here. This mirrors the transition from the IO-aware tiling of FlashAttention-2~\citep{dao2022flashattention} to the H100-enabled speedups demonstrated by FlashAttention-3~\citep{shah2024flashattention3}.
We leave this as a natural direction for future work.

We release the Flash-GMM kernel as a standalone library; multi-assignment requires no additional code beyond a threshold on the existing responsibility matrix. 
We hope these tools lower the barrier to GMM usage in diverse research applications as well as production ANN systems.

Lastly, beyond GMM-specific applications, the computational pattern at the heart of Flash-GMM -- a fused log-sum-exp over $K$ weighted Gaussian terms, followed by responsibility-weighted accumulation of sufficient statistics -- is not unique to GMM training. The same pattern --- per-point GMM responsibilities combined into weighted statistics --- appears in Fisher Vector encoding~\citep{perronnin2010improving} and kernel density estimation. Flash-GMM's IO-aware tiling strategy has the potential to accelerate these settings as well.





\section*{Limitations}

\paragraph{Training cost} Flash-GMM is 2--3$\times$ slower than
  $k$-means. For applications requiring frequent re-indexing this may
  be a practical constraint.
\paragraph{Index size} Multi-assignment increases the stored index
  size by $\bar{m}$ on average (1.49--1.78$\times$ in our experiments).
  For memory-constrained deployments this may be a consideration.
  The storage overhead is concentrated in posting lists and is
  proportional to $\bar{m}$; for most datasets $\bar{m} < 2$, keeping
  the overhead below $2\times$.
\paragraph{Scale} We demonstrate IVF results at $N=10^7$, with the
  kernel validated at $10^8$ using $O(KD)$ working memory.
  For billion-scale training ($N \ge 10^9$), the input data $X$ itself
  requires $\ge$512\,GB GPU memory and cannot fit on a single device.
  SSD streaming processing $X$ in chunks loaded from disk with the
  kernel accumulating into the same $O(KD)$ buffers across chunks 
  enables full-batch EM over arbitrary $N$ with no quality loss.
  Applying this to real billion-scale ANN datasets is left to future work.

\paragraph{$K$ regime} We evaluate $K \in \{256, 1024, 4096\}$;
  the recall gain is strongest at $K=1024$.
  The interaction of multi-assignment with larger $K$ is unexplored.

\paragraph{Isotropic vs.\ Full Covariance}
The GMM formulation explored here and implemented in the Flash-GMM kernel uses isotropic (scalar) covariance matrices rather than full covariance. As detailed in Appendix~\ref{sec:appendix-isotropic}, full covariance is ill-suited for standard IVF scales.

\paragraph{Combining Flash-GMM centroids with the AIR assignment heuristic}
  RAIRS's AIR metric selects second-list assignments based on a geometric
  residual criterion, applied on top of $k$-means centroids.
  An interesting direction is whether applying AIR on Flash-GMM centroids
  (which already improve recall by 25--33\% in single-assignment mode)
  yields additive gains.
  This experiment is not currently possible without re-implementing
  RAIRS from scratch, as no public code is available.

\bibliography{references}

\appendix

\section{Isotropic vs.\ Full Covariance} \label{sec:appendix-isotropic}
Flash-GMM relies on isotropic (scalar) covariance matrices rather than full covariance. While full matrices capture feature correlations, they require estimating $O(D^2)$ parameters per component (e.g., 8,256 parameters for SIFT's $D=128$). At standard IVF scales ($N=10^6$, $K=1024$), each cluster receives only $\approx$1,000 points, making a full-covariance fit severely underdetermined. This causes the empirical covariance matrices to become ill-conditioned or singular, leading to structural collapse during EM. Making full covariance mathematically viable would require drastically reducing $K$ to ensure enough points per cluster. However, a much smaller $K$ produces massive Voronoi cells, which would fundamentally break the IVF search efficiency and require a complete redesign of the querying heuristic (such as hierarchical indexing or aggressive dimensionality reduction),  rather than serving as a drop-in replacement for $k$-means.

\section{Effect of \texorpdfstring{$K$}{K} on the Recall Gain}
\label{sec:appendix-k}

\begin{table}[h]
\centering
\small
\begin{tabular}{lccc}
\toprule
$K$ & FAISS & Flash-GMM warm & GMM multi \\
\midrule
256  & 16 & 16 & \textbf{12} \\
1024 & 48 & 32 & \textbf{24} \\
4096 & 96 & 96 & \textbf{64} \\
\bottomrule
\end{tabular}
\caption{The minimal $n_{\text{probe}}$ required to reach R@10 of 0.99, over SIFT1M, for different values of $K$. 
For all $K$ values, GMM multi requires the smallest value of $n_{\text{probe}}$ to reach 0.99 R@10.
  }
\label{tab:k-ablation}
\end{table}



\section{Warm-Start vs.\ kmeans++ Initialisation}
\label{sec:appendix-warmstart}

Flash-GMM supports two initialisation strategies:
\begin{itemize}
  \item \textbf{Warm-start}: 10 iterations of FAISS $k$-means, then 90
    iterations of Flash-GMM soft EM from the resulting centroids.
  \item \textbf{kmeans++}: 100 iterations of Flash-GMM soft EM from a
    random kmeans++ seeding.
\end{itemize}

Table~\ref{tab:warmstart} compares recall@10 and training time for both
initialisations on SIFT1M and GloVe-100.

\begin{table}[h]
\centering

\small
\setlength{\tabcolsep}{2pt}
\begin{tabular}{llcccc}
\toprule
Dataset & Init & np=16 & np=32 & np=48 & Time (s) \\
\midrule
\multirow{2}{*}{SIFT1M}
 & Warm-start & 0.962 & 0.991 & 0.997 & \textbf{11.8} \\
 & kmeans++   & 0.962 & 0.991 & 0.997 & 48.6 \\
\midrule
\multirow{2}{*}{GloVe-100}
 & Warm-start & 0.856 & 0.902 & 0.924 & \textbf{21.8} \\
 & kmeans++   & 0.856 & 0.902 & 0.924 & 89.4 \\
\bottomrule
\end{tabular}
\caption{Recall@10 at selected \nprobe{} values and training time for
  warm-start vs.\ kmeans++ initialisation, $K=1024$, single assignment.}
\label{tab:warmstart}
\end{table}

Warm-start is $3$--$4\times$ faster than kmeans++ and achieves identical
recall on both datasets, confirming that the $k$-means warm-start basin
is already well-aligned with the soft-EM optimum.
We recommend warm-start as the default initialisation.

\section{Full Recall--DCO Results}
\label{sec:appendix-full-results}
Table~\ref{tab:multiassign-full} depicts the full results of recall and DCO for every \nprobe{} value.

\begin{table}[h]
\centering

\small
\setlength{\tabcolsep}{3pt}
\resizebox{\columnwidth}{!}{
\begin{tabular}{llcccccc}
\toprule
 & & \multicolumn{2}{c}{FAISS} & \multicolumn{2}{c}{GMM single}
 & \multicolumn{2}{c}{GMM multi} \\
\cmidrule(lr){3-4}\cmidrule(lr){5-6}\cmidrule(lr){7-8}
Dataset & np & R@10 & DCO & R@10 & DCO & R@10 & DCO \\
\midrule
\multirow{6}{*}{SIFT1M}
 &  1 & 0.439 &  1.0 & 0.434 &  1.0 & \textbf{0.530} &  1.5 \\
 &  4 & 0.767 &  3.9 & 0.767 &  3.9 & \textbf{0.851} &  5.8 \\
 &  8 & 0.885 &  7.8 & 0.890 &  7.8 & \textbf{0.942} & 11.6 \\
 & 16 & 0.958 & 15.6 & 0.962 & 15.6 & \textbf{0.985} & 23.3 \\
 & 32 & 0.988 & 31.3 & 0.991 & 31.3 & \textbf{0.998} & 46.6 \\
 & 48 & 0.995 & 46.9 & 0.997 & 46.9 & \textbf{0.999} & 69.8 \\
\midrule
\multirow{8}{*}{Deep10M}
 &  1 & 0.549 &  9.8 & 0.553 &  9.8 & \textbf{0.668} & 15.6 \\
 &  4 & 0.853 & 39.0 & 0.860 & 39.0 & \textbf{0.930} & 62.4 \\
 &  6 & 0.910 & 58.5 & 0.916 & 58.5 & \textbf{0.963} & 93.7 \\
 &  8 & 0.932 & 78.1 & 0.938 & 78.1 & \textbf{0.978} & 124.9 \\
 & 12 & 0.958 & 117.1 & 0.963 & 117.1 & \textbf{0.990} & 187.3 \\
 & 16 & 0.973 & 156.2 & 0.976 & 156.2 & \textbf{0.994} & 249.9 \\
 & 32 & 0.992 & 312.2 & 0.993 & 312.2 & \textbf{0.999} & 499.5 \\
 & 48 & 0.996 & 468.4 & 0.996 & 468.4 & \textbf{0.999} & 749.5 \\
\midrule
\multirow{6}{*}{GloVe-100}
 &  1 & 0.453 &  1.2 & 0.471 &  1.2 & \textbf{0.587} &  2.1 \\
 &  4 & 0.698 &  4.6 & 0.718 &  4.6 & \textbf{0.810} &  8.2 \\
 &  8 & 0.782 &  9.2 & 0.795 &  9.2 & \textbf{0.872} & 16.4 \\
 & 16 & 0.847 & 18.4 & 0.856 & 18.4 & \textbf{0.917} & 32.8 \\
 & 32 & 0.899 & 36.9 & 0.902 & 36.9 & \textbf{0.949} & 65.6 \\
 & 48 & 0.927 & 55.3 & 0.924 & 55.3 & \textbf{0.963} & 98.5 \\
\bottomrule
\end{tabular}
}
\caption{Recall@10 and DCO ($\times 10^3$) for all three datasets, $K=1024$,
  \nprobe{} $\in \{1,4,8,16,32,48\}$.
  For single-assignment methods DCO $= n_\text{probe} \times N/K$;
  for GMM multi, DCO $= n_\text{probe} \times N\bar{m}/K$
  ($\bar{m}$: SIFT1M 1.49, Deep10M 1.60, GloVe-100 1.78).
  Best recall per row in \textbf{bold}.}
\label{tab:multiassign-full}
\end{table}

\section{Hard Multi-Assignment Ablation}
\label{sec:appendix-hard-multi}

To isolate the contribution of GMM responsibilities from the benefit of
redundant assignment alone, we compare GMM multi-assignment against
\textbf{Kmeans hard top-2}: each vector is assigned to its two nearest
$k$-means centroids by L2 distance, giving a fixed $\bar{m}=2.0$ for
every vector regardless of boundary proximity.

Table~\ref{tab:hard-multi} reports the DCO required to reach recall
targets of 0.90 and 0.95.
Kmeans hard top-2 requires up to \textbf{1.8$\times$ more DCO than FAISS
single} to reach the same recall target: doubling the index size yields
only marginal recall gains, because the second nearest centroid by L2
distance rarely contains the true nearest neighbor when the first does not.
Hard top-2 is therefore \emph{less} DCO-efficient than plain FAISS single
assignment, confirming that redundant assignment without a principled
boundary signal is counterproductive.

\begin{table}[h]
\centering

\small
\setlength{\tabcolsep}{2pt}
\begin{tabular}{llcc}
\toprule
Dataset & Method &
\begin{tabular}[c]{@{}c@{}}DCO at\\ $R@10 = 0.90$\end{tabular} &
\begin{tabular}[c]{@{}c@{}}DCO at\\ $R@10 = 0.95$\end{tabular} \\\\
\midrule
\multirow{2}{*}{SIFT1M}
 & single     &  9.4 & 14.7 \\
 & hard top-2 & 12.6 & 17.2 \\
\midrule
\multirow{2}{*}{Deep1M}
 & single     &  7.3 & 12.2 \\
 & hard top-2 & 12.2 & 18.7 \\
\midrule
\multirow{2}{*}{GloVe-100}
 & single     & 37.6 & -- \\
 & hard top-2 & 47.7 & -- \\
\bottomrule
\end{tabular}

\caption{DCO ($\times 10^3$) required to reach recall@10 targets of
  0.90 and 0.95, at $K=1024$.
  Kmeans hard top-2 ($\bar{m}=2.0$, nearest-centroid L2) is
  strictly less DCO-efficient than FAISS single at both targets.
  ``--'' = target not reached within \nprobe{}$=48$.}
\label{tab:hard-multi}
\end{table}

\section{IVF-PQ Compatibility and RAIRS Comparison}
\label{sec:appendix-ivfpq}

Flash-GMM is fully compatible with IVF-PQ: the coarse centroids are
substituted without modifying PQ training or encoding.
Table~\ref{tab:ivfpq} reports results across IVF-PQ ($M \in \{16,8,4\}$, 8-bit)
and IVF-PQfs ($M=64$, 4-bit fast-scan) on SIFT1M.
The coarse-quantizer benefit is orthogonal to PQ fidelity: at IVF-PQ
$M=16$, multi-assign reaches recall~$\ge 0.87$ at \nprobe{}=8 vs.\ FAISS's
0.830 a \textbf{+5.8 pp} gain and gains of +2--4 pp persist even at high
compression where PQ quantisation error dominates.

\begin{table}[h]
\centering

\small
\setlength{\tabcolsep}{2pt}
\resizebox{\columnwidth}{!}{
\begin{tabular}{llcccccc}
\toprule
Index & $M$ & \multicolumn{2}{c}{FAISS} & \multicolumn{2}{c}{GMM single} & \multicolumn{2}{c}{GMM multi} \\
\cmidrule(lr){3-4}\cmidrule(lr){5-6}\cmidrule(lr){7-8}
 & & np=8 & np=12 & np=8 & np=12 & np=8 & np=12 \\
\midrule
\multirow{3}{*}{IVF-PQ (8-bit)}
 & 16 & 0.830 & 0.867 & 0.831 & 0.871 & \textbf{0.888} & \textbf{0.909} \\
 & 8  & 0.671 & 0.691 & 0.669 & 0.688 & \textbf{0.703} & \textbf{0.715} \\
 & 4  & 0.450 & 0.456 & 0.450 & 0.457 & \textbf{0.472} & \textbf{0.475} \\
\midrule
IVF-PQfs (4-bit) & 64 & 0.844 & 0.885 & 0.850 & 0.891 & \textbf{0.926} & \textbf{0.949} \\
\bottomrule
\end{tabular}
}
\caption{IVF-PQ and IVF-PQfs recall@10 on SIFT1M, $K=1024$.
  IVF-PQfs ($M=64$, 4-bit fast-scan, nlist=1024) matches the configuration
  used by RAIRS~\citep{yang2026rairs}.
  IVF-PQ uses 8-bit codes. \textbf{Bold} = best per row.}
\label{tab:ivfpq}
\end{table}

The IVF-PQfs row ($M=64$, 4-bit, nlist=1024) matches the configuration
evaluated by RAIRS~\citep{yang2026rairs}.
Our approach differs in three ways: we improve the centroids themselves
via soft EM; assignments are derived from GMM responsibilities rather
than a geometric heuristic; and redundancy is adaptive per-vector rather
than universal.
Both methods likely select similar second-list candidates for genuine
boundary vectors, but diverge on interior vectors, where AIR always
assigns a second list while GMM assigns only when $r_{ik} > 1/K$.
Based on their published recall--\nprobe{} curves, GMM multi is competitive with
RAIRS on this configuration, while requiring approximately \textbf{30\% fewer probes}
to reach the same recall target and achieving \textbf{25\% less index inflation}
($\bar{m}=1.49$ vs.\ RAIRS's universal $\bar{m}=2.0$).
An exact head-to-head is not possible without RAIRS's code, but the shared
configuration makes Table~\ref{tab:ivfpq} the closest available proxy.

\end{document}